\documentclass[conference]{IEEEtran}

\IEEEoverridecommandlockouts

\usepackage{booktabs}
\usepackage{caption}
\usepackage{rotating}
\usepackage{xcolor}
\usepackage{soul}
\usepackage{hyperref}

\usepackage{placeins} 

\def\BibTeX{{\rm B\kern-.05em{\sc i\kern-.025em b}\kern-.08em
    T\kern-.1667em\lower.7ex\hbox{E}\kern-.125emX}}

\begin{document}

\title{The RaspGrade Dataset: Towards Automatic Raspberry Ripeness Grading with Deep Learning \\

}

\author{\IEEEauthorblockN{Mohamed Lamine Mekhalfi}
\IEEEauthorblockA{\textit{Fondazione Bruno Kessler, Trento, Italy} \\
mmekhalfi@fbk.eu}
\and

\IEEEauthorblockN{Paul Chippendale}
\IEEEauthorblockA{\textit{Fondazione Bruno Kessler, Trento, Italy} \\
chippendale@fbk.eu}
\and

\IEEEauthorblockN{Fabio Poiesi}
\IEEEauthorblockA{\textit{Fondazione Bruno Kessler, Trento, Italy} \\
poiesi@fbk.eu}
\and

\IEEEauthorblockN{Samuele Bonecher}
\IEEEauthorblockA{\textit{Sant'Orsola, Pergine Valsugana, Italy} \\
samuele.bonecher@santorsola.com}
\and

\IEEEauthorblockN{Gilberto Osler}
\IEEEauthorblockA{\textit{Sant'Orsola, Pergine Valsugana, Italy} \\
gilberto.osler@santorsola.com}
\and

\IEEEauthorblockN{Nicola Zancanella}
\IEEEauthorblockA{\textit{Sant'Orsola, Pergine Valsugana, Italy} \\
nicola.zancanella@santorsola.com}
\and
}

\maketitle

\begin{abstract}
This research investigates the application of computer vision for rapid, accurate, and non-invasive food quality assessment, focusing on the novel challenge of real-time raspberry grading into five distinct classes within an industrial environment as the fruits move along a conveyor belt. To address this, a dedicated dataset of raspberries, namely  \textit{RaspGrade}, was acquired and meticulously annotated. Instance segmentation experiments revealed that accurate fruit-level masks can be obtained; however, the classification of certain raspberry grades presents challenges due to color similarities and occlusion, while others are more readily distinguishable based on color. The acquired and annotated \textit{RaspGrade} dataset is accessible on Hugging Face at: \href{https://huggingface.co/datasets/FBK-TeV/RaspGrade}{https://huggingface.co/datasets/FBK-TeV/RaspGrade}.

\end{abstract}


\section{Introduction}
Machine vision offers a non-invasive, customizable, and cost-effective technological advancement in fruit and vegetable inspection, providing a sophisticated means of optimizing both efficiency and accuracy, resulting in increased throughput.  This technology finds application in a variety of tasks, ranging from yield estimation and defect detection to quality grading and packaging inspection, and is suitable for deployment in both outdoor and indoor environments \cite{aiadi2022date, mamdouh2021yolo, gao2022novel}.

Yet, real-time raspberry fruit grading in an industrial setting is a challenging but interesting problem that, to the best of our knowledge, has not been addressed in prior literature. Its challenge stems from the fact that raspberries are small fruits that can be difficult to detect, besides the similarity in color among raspberry grades.
Sant’Orsola is an Italian leader organization of producers specialized in the cultivation of small fruits (e.g.,~strawberries, raspberries, cherries). 
The company aims to integrate automated solutions to improve the quality of its products and foster its throughput. 
In this context, a key task targeted for automation is the grading of raspberries.  This process involves the identification of each fruit's grade within a transparent punnet as it traverses the packaging line's conveyor belt.  Currently, this grading procedure is executed manually by human operators immediately preceding packaging.  This allows experts to remove any fruit that fails to meet established quality standards.

Extensive research has contributed significantly to the domain of small fruit segmentation and analysis \cite{chen2023instance, ni2020deep, zabawa2020counting, akiva2020finding, riz2024wild, perez2020fast, buayai2020end, nuske2011yield}. Nevertheless, a significant portion of this existing body of work focuses on acquisition scenarios distinct from industrial settings, with many studies addressing fruit detection in field environments prior to harvesting. Furthermore, to the best of our knowledge, the challenge of raspberry grading based on ripeness remains an open problem, and no publicly available dataset has been presented to date. This research endeavors to address these identified limitations.
In particular, the objective of this study is to develop a vision system capable of classifying raspberries into five distinct grades based on ripeness, effectively automating the process currently performed manually by human experts.
Specifically, the envisioned classification encompasses five grades: OK (Grade 1), Dark (Grade 2), Light (Grade 3), Second (Grade 4), and Waste (Grade 5). To address this need, this research aims to answer the following question: To what extent can a deep learning-based instance segmentation model (YOLOv8) accurately classify raspberries into five distinct ripeness grades in a simulated real-time industrial setting using RGB imagery? To answer this question, this paper contributes with the following:


\begin{itemize}
    \item We address the novel problem of real-time raspberry fruit grading based on ripeness in an industrial setting, a task that has not been previously addressed in the literature.
    \item We collect a dataset (\textit{RaspGrade}) of RGB images of raspberry fruits, covering five different ripeness grades, including pixel-level segmentation masks for each fruit and class labels indicating ripeness level.  This dataset will be released publicly, which we believe will be a valuable resource for the community, enabling further research on automated fruit grading.
    \item We conduct a quantitative experimental methodology and provide baseline results, suggesting the viability of the proposed approach in real-time scenarios.
\end{itemize}

This paper is organized as follows. Section \ref{related} reviews existing art. Section \ref{methodology} describes the methodology. Section \ref{experiments} conducts the experiments and results discussion. Section \ref{conclusions} concludes the paper.

\section{Related works} \label{related}

The relevant literature has accumulated interesting contributions for fruit and vegetable inspection. 
For instance, 
Mahmud et al. \cite{mahmud2020real} developed a system for detecting powdery mildew on strawberries using a mobile mini-vehicle.  This vehicle was equipped with dual optical sensors for wide-field image capture, a GPS module for mission planning, a laptop for on-board processing, and an artificial lighting system (using a black cloth enclosure) to mitigate the effects of direct sunlight.  A serial link facilitated communication between the GPS and the laptop.  Feature extraction was performed using the color co-occurrence matrix, and these features were subsequently used to train an artificial neural network for disease classification.
Yang et al. \cite{yang2023automatic} tackled the challenge of potato defect detection by combining multispectral imaging with deep learning techniques.  Their approach utilized 25 wavebands to capture detailed information about the potatoes and addressed the identification of five distinct defect types: germination, common scab, bug-eye, dry-rot, and bruise.  This method achieved a mean average precision of 90.26\%.
The detection of on-tree plum fruits was explored in \cite{tang2023yolov7}, where an adapted version of the state-of-the-art deep learning model, namely YOLOv7, was utilized on high-resolution images of plum fruits, yielding promising results. Postharvest storage plays a critical role in the fruit supply chain, as factors like humidity, temperature, and ventilation can significantly impact fruit quality. Maintaining optimal storage conditions helps reduce losses and ensures a continuous supply. Additionally, pre-storage fruit analysis can assist in detecting early-stage diseases.
Wan et al. \cite{wan2018methodology} investigated the use of color features for determining the maturity of Roma and Pear tomato varieties.  While traditional machine vision systems often rely on visible light sensors for assessing produce characteristics such as size, count, and surface defects, these systems are limited in their ability to detect a range of plant diseases, many of which occur outside the visible spectrum.  Multispectral imaging offers a solution to this limitation.  For instance, De et al. (2023) employed multispectral imaging in conjunction with convolutional neural networks and vision transformers to detect diseases in tomato, potato, and papaya leaves.
Leiva et al. \cite{leiva2013automatic} investigated the application of computer vision to the classification of blueberry diseases.  Their approach involved segmenting individual blueberries from acquired images and subsequently extracting a range of features, including textural, intensity, and geometric properties.  The effectiveness of several classification methods was then evaluated, with Support Vector Machines and Linear Discriminant Analysis demonstrating the highest performance.
Unay et al. \cite{unay2022deep} explored the use of multispectral imaging and deep learning for apple quality grading. While single-modality imaging can be sufficient for many machine vision tasks in fresh produce processing, integrating multiple modalities offers advantages for more complex applications.  Wang et al. \cite{wang2015multimodal}, for example, combined hyperspectral, 3D, and X-ray imaging for onion quality inspection, achieving an 88.9\% classification score in distinguishing healthy from defective onions.  However, this approach predates the widespread adoption of advanced deep learning techniques, which have the potential to significantly improve performance.  Mesa et al. \cite{mesa2021multi} demonstrated this potential by using a multimodal approach with RGB and hyperspectral imaging for banana grading into three classes.  Their deep learning model, trained on this combined data, achieved a remarkable 98.45\% overall accuracy.

Accurate yield estimation is crucial for effective logistical planning in agriculture.  Overestimating yields can lead to unnecessary expenditures on storage and transportation, while underestimation can result in critical shortages later in the supply chain.  Mekhalfi et al. \cite{mekhalfi2020vision} addressed the problem of kiwi fruit yield estimation through a vision-based approach.  Their system employed an optical sensor mounted on a slow-moving tractor to survey orchards.  The captured images were then processed by software that performed image pre-processing, stitching, and fruit counting to provide a yield estimate.  To overcome the challenge of foliage obstructing sunlight and hindering fruit detection, the system incorporated an upward-facing LED light source positioned adjacent to the camera, illuminating the fruit for optimal visibility.
Gongal et al. \cite{gongal2016apple} developed an over-the-row machine vision system for apple fruit counting and yield estimation.  Their system consisted of a tunnel-like enclosure, RGB 3D sensors, and LED lighting.  A key aspect of their work was a comparison between single-side imaging and dual imaging, where both sides of the apple trees were captured.  The dual-imaging approach yielded significantly higher crop estimation accuracy 82\% compared to the single-side method 58\%.  The use of the enclosed housing and controlled LED lighting provided the benefit of consistent illumination, making the system independent of natural light conditions and enabling operation both day and night.

For an in-depth analysis, the work presented in \cite{vasudevan2024robotics} surveys various machine vision systems and their components for food item analysis.  

\section{Methodology} \label{methodology}
\subsubsection{Problem statement}
\begin{figure*}

\includegraphics[scale=0.32]{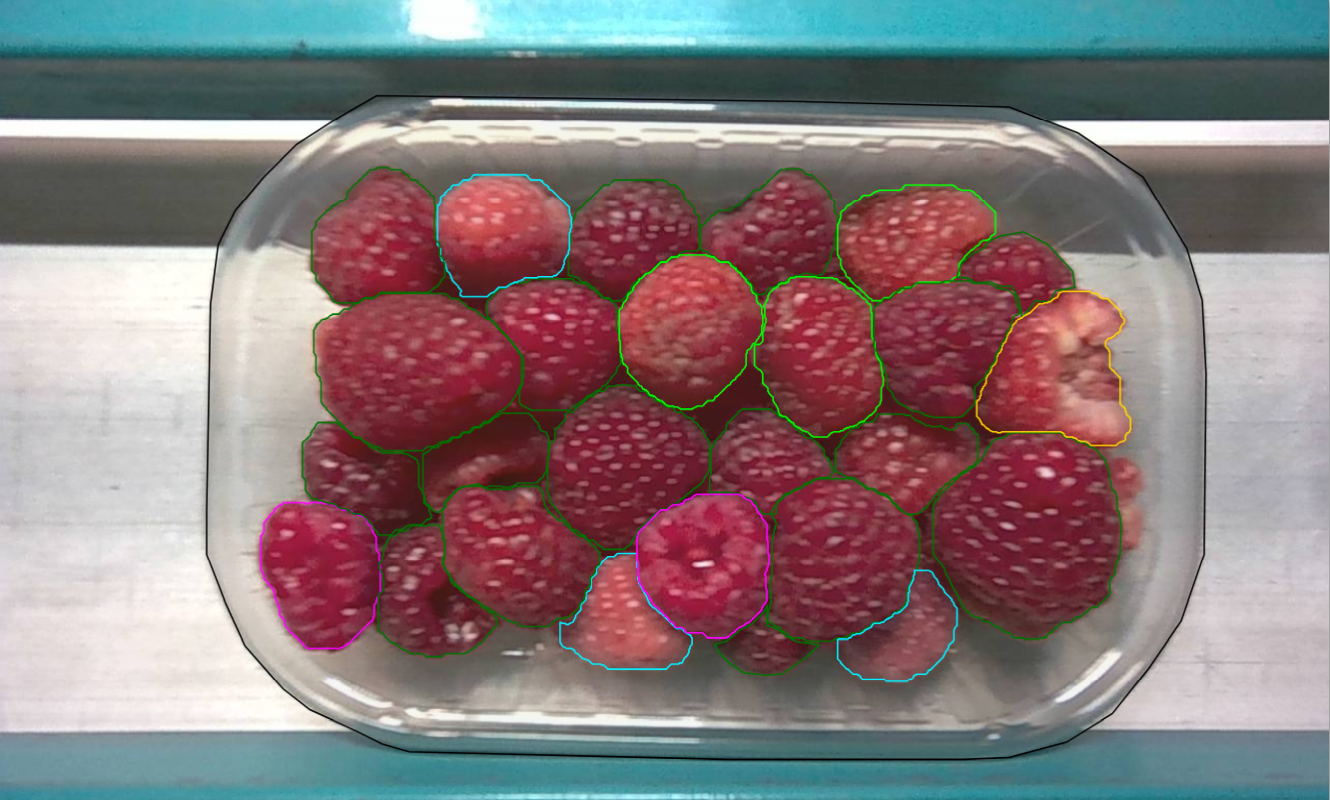}
\includegraphics[scale=0.32]{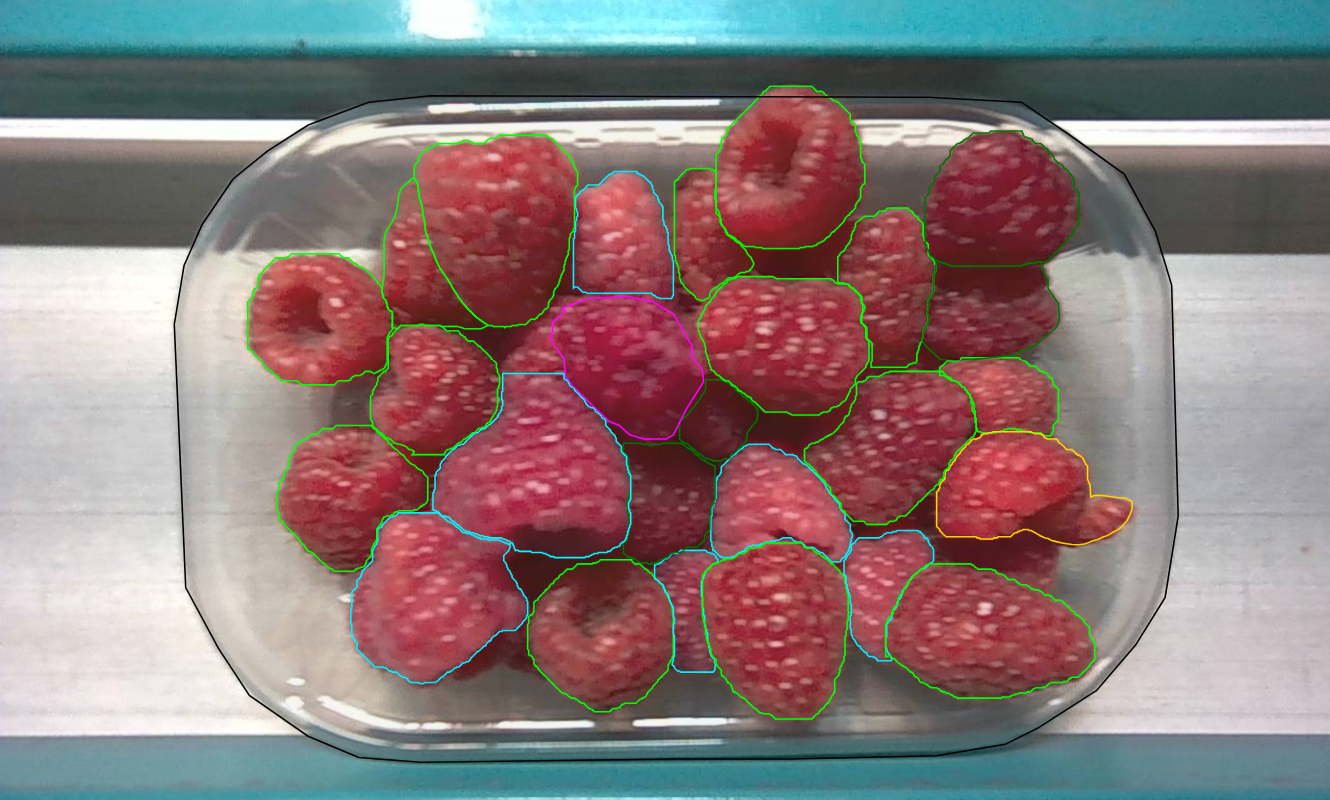}

\caption{Raspberry and punnet annotation examples using CVAT \cite{cvat}. Light green annotations: Grade 1. Dark green: Grade 2. Cyan: Grade 3. Orange: Grade 4. Purple: Grade 5. Black: punnet.}
\label{fig_grades}
\end{figure*}

As previously mentioned, grading into five classes is based on fruit ripeness, which is directly correlated with raspberry color. To illustrate this, Fig.~\ref{fig_grades} presents two sample punnets containing various raspberry classes that were annotated via the CVAT tool \cite{cvat}.

The removal of Waste grade fruit, while outside the scope of this paper, is addressed using a robotic arm. This arm receives the coordinates and grade of each fruit within the punnet, along with the punnet's own coordinates, to facilitate pick-and-place operations. The coordinate of each fruit or punnet is defined as the centroid pixel of all pixels belonging to the segment that defines the fruit or punnet.  To this end, the grading task constitutes an instance segmentation problem, where, for each fruit and punnet, both the constituent pixels and the corresponding class label are inferred simultaneously.
Several challenges characterize this instance segmentation task. These include: (i) the requirement for real-time grading as fruit punnets move along the conveyor belt, which can lead to image blurring, thus compromising the edges and texture defining individual raspberries and punnets; (ii) the small size of raspberries, which may result in insufficient pixel coverage for accurate segmentation; (iii) the significant impact of image acquisition viewpoint on system performance, as individual fruits can exhibit varying profiles from different angles; and (iv) the frequent occlusion of individual raspberries, which complicates the segmentation process.

While several instance segmentation models exist in the literature \cite{charisis2024deep}, including two-stage detectors like Mask R-CNN \cite{he2017mask} and transformer-based architectures such as DETR \cite{carion2020end}, we opted for the YOLOv8 model \cite{yolov8}. Our choice was primarily driven by the critical requirement for real-time performance in the industrial setting of Sant'Orsola's packaging line. YOLOv8, as a state-of-the-art one-stage detector, is renowned for its efficiency and high inference speeds, making it more suitable for real-time applications compared to the potentially higher latency of two-stage methods. Furthermore, YOLOv8 achieves a compelling balance between speed and accuracy in object detection and instance segmentation tasks. While transformer-based models have shown promising results in object detection, their computational cost can be higher, which might be a limiting factor for real-time deployment. The one-stage design of YOLOv8 allows for direct prediction of instance masks, contributing to its efficiency. 

\subsubsection{Data acquisition}
\begin{figure*}
    \centering
    \begin{tabular}{@{}cc@{}} 
    
        \includegraphics[width=0.45\textwidth]{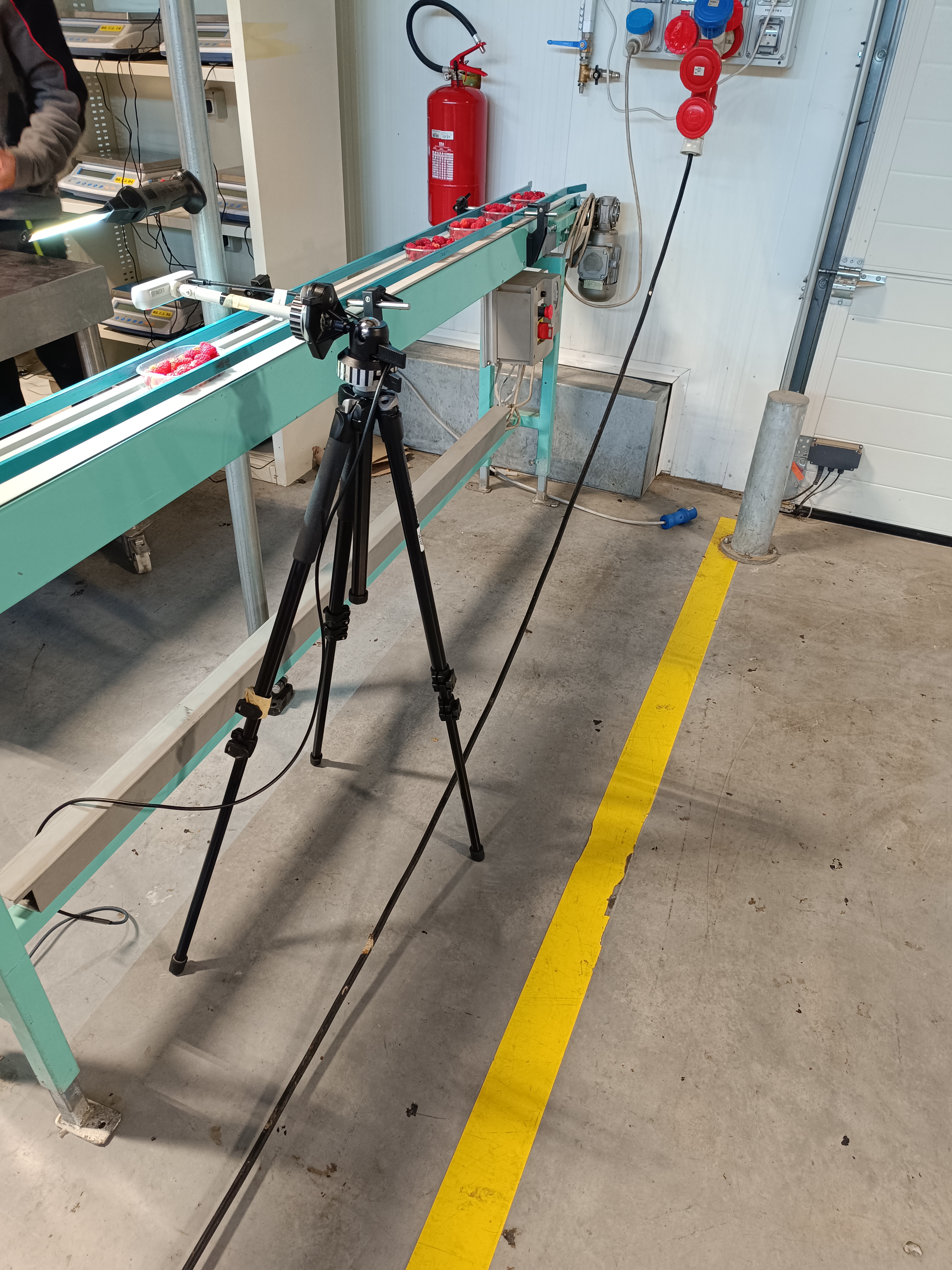}
        \includegraphics[width=0.45\textwidth]{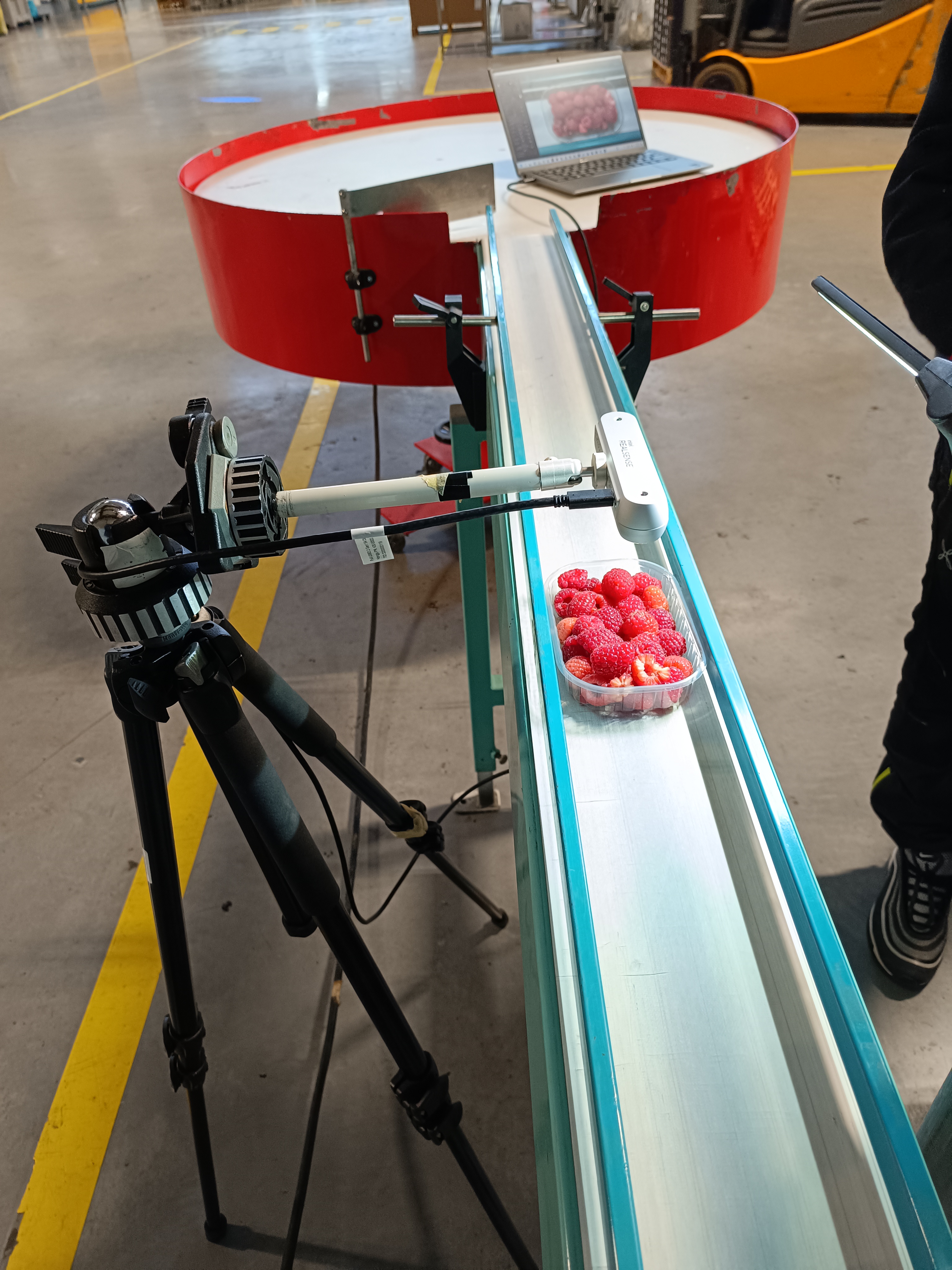} \\
        
        \includegraphics[width=0.45\textwidth]{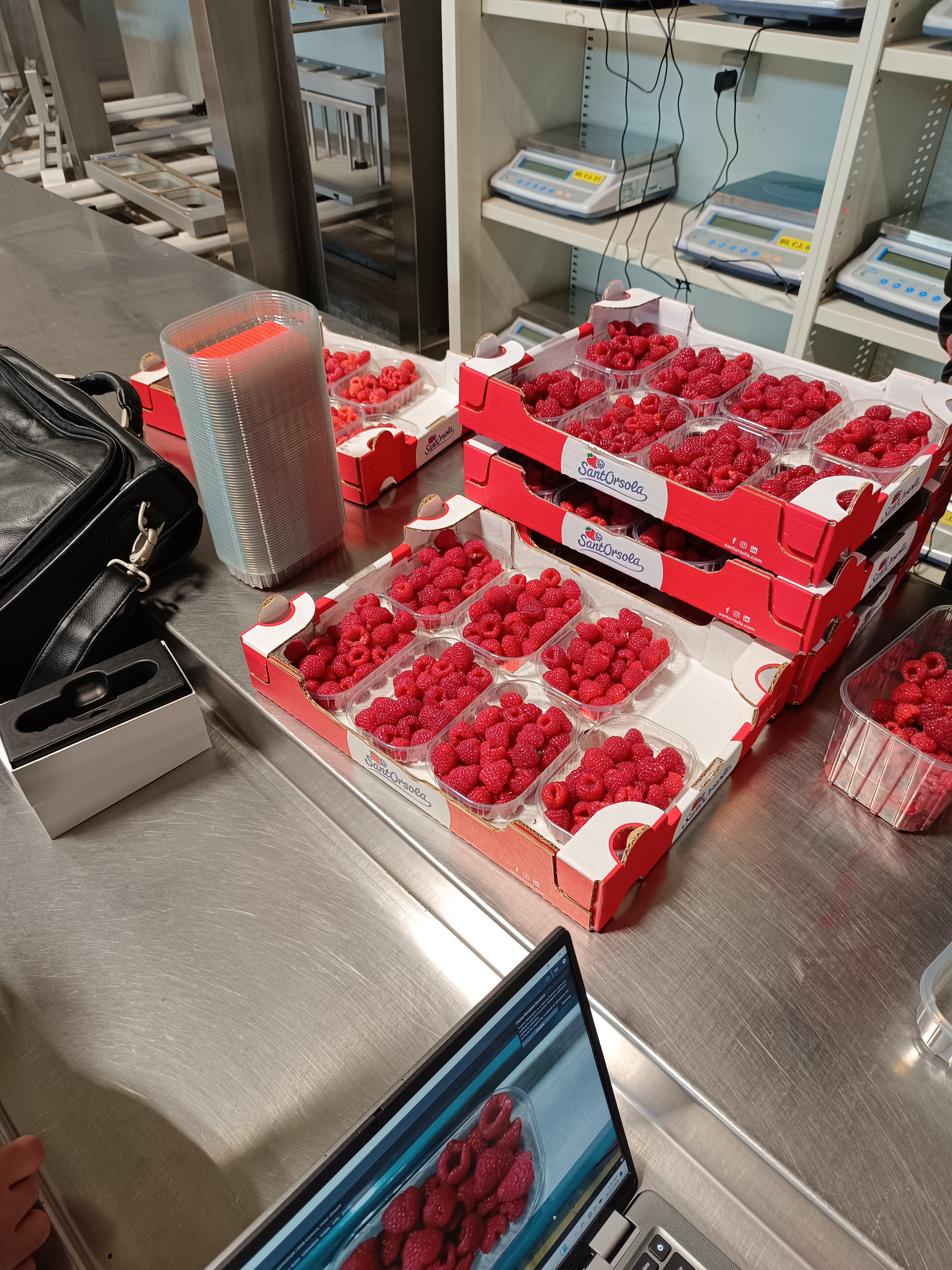}
        \includegraphics[width=0.45\textwidth]{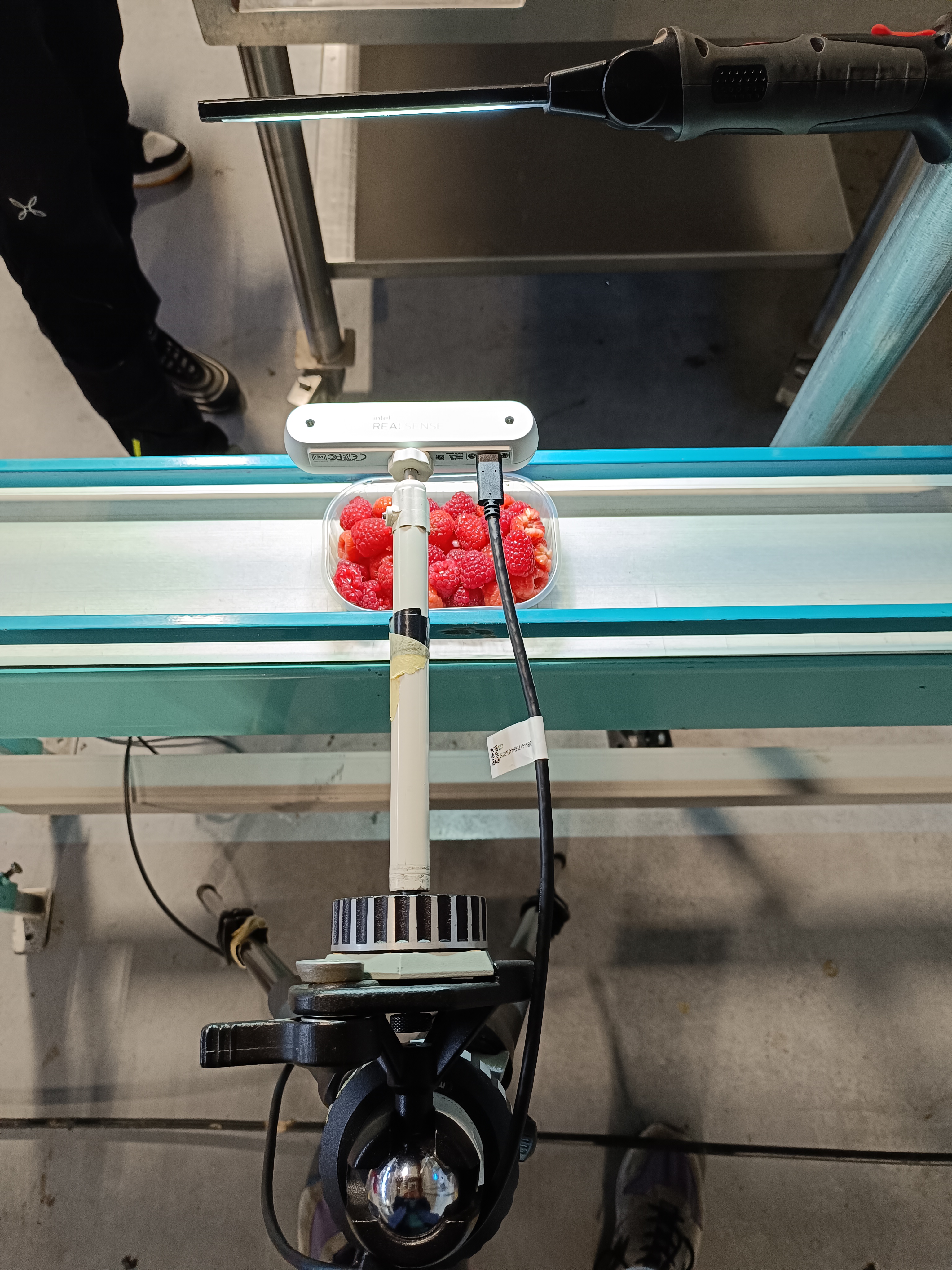} \\

    \end{tabular}
    
\caption{Acquisition setup viewed from different angles.}
\label{fig_acquisition}
    
\end{figure*}

The \textit{RaspGrade} dataset of RGB images of raspberries and punnets was acquired. 
The acquisition setup, illustrated in Fig.~\ref{fig_acquisition}, consists of a tripod supporting a vertical camera mount. At the end of this mount, an Intel® RealSense™ D456 sensor \cite{realsense} is positioned to capture a top-down view. Slightly above the camera level, we fixed a LED lighting (ERGOPOWER PEN+) from WÜRTH \cite{wurth} in order to illuminate the punnet and the raspberries therein.

The collected images were manually annotated using  CVAT annotation tool \cite{cvat}. This consists of annotating the segment as well as the grade of each fruit (this latter was performed by experts from Sant’Orsola). 

The statistics of the acquired data are given in Table.~\ref{tab:stats}. In total, 200 punnets containing 5243 raspberry fruits were acquired (on average, roughly 26 fruits are envisioned per punnet), which were split into 4202 instances dedicated for the training of the network (accounting for 80\%), and the remaining 1041 were allocated for testing purposes (accounting for 20\%). The samples belonging to Grade 2 constitute the majority while the Second and the Waste categories account for the smallest subsets, and the effect of this will be discussed in the next section.

\begin{table}
\captionsetup{size=footnotesize}
\caption{Dataset statistics.}
\setlength\tabcolsep{3pt} 
\centering
\begin{tabular}{lcccccccc}
\toprule

\textbf{Class} & Training & Validation & Total\\

\midrule  
\textbf{Punnet} & 160 & 40 & 200 \\
\textbf{OK (Grade 1)} & 773 & 205 & 978 \\
\textbf{Dark (Grade 2)} & 2879 & 724 & 3603 \\
\textbf{Light (Grade 3)} & 262 & 44 & 306 \\
\textbf{Second (Grade 4)} & 50 & 12 & 62 \\
\textbf{Waste (Grade 5)} & 78 & 16 & 94 \\
\midrule
\textbf{All} & 4202 & 1041 & 5243 \\

\bottomrule
\end{tabular}
\label{tab:stats}
\end{table}
\normalsize

\section{Experiments} \label{experiments}

\subsection{Implementation details}
We train the YOLOv8 model for 100 epochs using the $yolov8n-seg$ backbone with a batch size of 2. For optimal quality in capturing fine details, such as the small features of raspberries, we select the highest image resolution of 1280$\times$800.
All the other parameters are set to default values as indicated in \cite{ultralytics}, unless otherwise specified. 
Regarding the classification component of the Loss function, we utilize the Binary Cross-Entropy Loss (BCE), and the per-class weights are explored in the next subsection. 

To evaluate the model's performance, we use standard metrics in object detection and instance segmentation tasks. Specifically:
\begin{itemize}
    \item Precision (Pre): measures the proportion of true positive detections out of all predicted detections, serving as an indicator of the accuracy of the model's predictions.
    \item Recall (Rec): quantifies the proportion of true positive detections out of all actual ground truth instances, highlighting how well the model captures all relevant objects.
    \item Mean Average Precision (mAP50): represents the mean of the average precision scores computed for each object class, summarizing the precision-recall curve by averaging precision at various recall levels.
    \item mAP50-95: A variant of mAP, it computes average precision across multiple Intersection over Union (IoU) thresholds, ranging from 0.50 to 0.95, with a step size of 0.05. This provides a comprehensive evaluation of the model's performance across different thresholds.
\end{itemize} 
We report the best scores obtained for each metric on the validation set during training. While the primary focus of this paper is on instance segmentation, we also present the raspberry detection scores, which may contribute to future research endeavors.
In the following tables, the segmentation and detection results are labeled as Mask and Box, respectively.

\subsection{Results and discussion}
We first report the results for each class based on the default parameters defined in \cite{ultralytics} in  Table.~\ref{table1}. An overall mAP of 62.20\% was obtained. We note that the Punnet class was segmented with a mAP of 99.5\%, which is rational given the large size of the punnets in the input images, and their consistent shape and color across the images of the dataset.

\begin{table*}
\captionsetup{size=footnotesize}
\caption{Raspberry instance segmentation and detection scores (\%) obtained with the default parameters. Box metrics stand for the detection task, and Mask metrics refer to the instance segmentation task.}
\setlength\tabcolsep{3pt} 
\centering
\smallskip 
\begin{tabular}{lcccccccc}
\toprule

\textbf{Class} & \textbf{Pre(Box)} & \textbf{Rec(Box)} & \textbf{mAP50(Box)} & \textbf{mAP50-95(Box)} & \textbf{Pre(Mask)} & \textbf{Rec(Mask)} & \textbf{mAP50(Mask)} & \textbf{mAP50-95(Mask)} \\

\midrule

\textbf{All} & 56.50 & 70.00 & 62.00 & 56.30 & 56.60 & 70.20 & 62.20 & 53.30 \\
\midrule  
\textbf{Punnet} & 97.50 & 100.0 & 99.50 & 93.00 & 97.50 & 100.0 & 99.50 & 77.70 \\
\textbf{OK (Grade 1)} & 58.40 & 71.20 & 70.70 & 63.20 & 58.40 & 71.20 & 71.00 & 63.80 \\
\textbf{Dark (Grade 2)} & 80.70 & 94.60 & 90.40 & 79.40 & 81.40 & 95.40 & 91.10 & 79.10 \\
\textbf{Light (Grade 3)} & 40.50 & 71.20 & 59.20 & 57.80 & 40.50 & 71.20 & 59.20 & 55.70 \\
\textbf{Second (Grade 4)} & 45.70 & 58.30 & 42.90 & 38.00 & 45.70 & 58.30 & 42.90 & 37.20 \\
\textbf{Waste (Grade 5)} & 16.00 & 25.00 & 9.30 & 6.40 & 16.00 & 25.00 & 9.30 & 6.30 \\

\bottomrule
\end{tabular}
\label{table1}
\end{table*}
\normalsize

As per the raspberries, Grade 2 scores the highest mAP of 91.10\%, owing mainly to the large number of training samples dedicated to this class (i.e.,~2879 samples among a total of 4202 training raspberry instances), followed by Grade 1 and Grade 3 raspberries, with mAPs of 71.00\% and 59.20\% respectively, which are also proportional to the number of samples in these two classes (respectively 773 and 262 instances, over a total of 4202). By contrast, although Grade 4 contributes with fewer samples than Grade 5 in the training set (50 versus 75 respectively, over a total of 4202), the segmentation score of Grade 4 is much higher than that of Grade 5 (42.90\% versus 9.30\%). This may be due to the fact that Grade 5 raspberries often manifest bruises and cuts on the surface, which are often occluded by other raspberries or show up on the invisible side of the fruit, which makes their detection difficult.

In quest of dealing with the Grade 5 bottleneck explained above, we increase the weight of this class in the aforementioned BCE loss to 10, and keep the weights of all the other classes to the default value of 1. The results are summarised in Table.~\ref{table2}. Indeed, the mAP of Grade 5 has increased from 9.30\% to 17.5\%, owing to the assigned high weight in the BCE classification loss, which forces the model to pay more attention to this class. However, this comes at the expense of Grade 3 and Grade 4 raspberries, whose mAPs have dropped to 53.60\% and 35.50\%, respectively. 

\begin{table*}
\captionsetup{size=footnotesize}
\caption{Raspberry instance segmentation and detection scores (\%) obtained by setting the Waste (Grade 5) loss weight to 10. Box metrics stand for the detection task, and Mask metrics refer to the instance segmentation task.}
\setlength\tabcolsep{3pt} 
\centering
\smallskip 
\begin{tabular}{lcccccccc}
\toprule

\textbf{Class} & \textbf{Pre(Box)} & \textbf{Rec(Box)} & \textbf{mAP50(Box)} & \textbf{mAP50-95(Box)} & \textbf{Pre(Mask)} & \textbf{Rec(Mask)} & \textbf{mAP50(Mask)} & \textbf{mAP50-95(Mask)} \\

\midrule

\textbf{All} & 59.50 & 67.10 & 61.50 & 57.00 & 59.60 & 67.20 & 61.60 & 53.40 \\
\midrule  
\textbf{Punnet} & 98.00 & 100.0 & 99.50 & 99.50 & 98.00 & 100.0 & 99.50 & 86.60 \\
\textbf{OK (Grade 1)} & 62.90 & 72.20 & 71.30 & 63.80 & 62.80 & 72.20 & 71.50 & 60.30 \\
\textbf{Dark (Grade 2)} & 78.40 & 96.50 & 91.60 & 79.00 & 78.70 & 97.10 & 91.80 & 77.10 \\
\textbf{Light (Grade 3)} & 40.70 & 77.30 & 53.60 & 48.50 & 40.60 & 77.30 & 53.60 & 48.50 \\
\textbf{Second (Grade 4)} & 55.80 & 31.80 & 35.50 & 35.20 & 55.90 & 31.90 & 35.50 & 32.90 \\
\textbf{Waste (Grade 5)} & 21.30 & 25.00 & 17.50 & 16.00 & 21.30 & 25.00 & 17.50 & 15.10 \\

\bottomrule
\end{tabular}
\label{table2}
\end{table*}
\normalsize

To improve the classification performance of the model, we increased the weight of the classification loss from a default 0.5 to 7, and the results are given in Table.~\ref{table3}. Interestingly, the overall mAP improved significantly to 65.50\%, thanks to the increase in the mAPs of Grade 1, Grade 3 and Grade 5 compared to the baseline results of Table.~\ref{table1}.

\begin{table*}
\captionsetup{size=footnotesize}
\caption{Raspberry instance segmentation and detection scores (\%) obtained by setting the Waste (Grade 5) loss weight to 10 and the YOLOv8 classification loss weight to 7. Box metrics stand for the detection task, and Mask metrics refer to the instance segmentation task.}
\setlength\tabcolsep{3pt} 
\centering
\smallskip 
\begin{tabular}{lcccccccc}
\toprule

\textbf{Class} & \textbf{Pre(Box)} & \textbf{Rec(Box)} & \textbf{mAP50(Box)} & \textbf{mAP50-95(Box)} & \textbf{Pre(Mask)} & \textbf{Rec(Mask)} & \textbf{mAP50(Mask)} & \textbf{mAP50-95(Mask)} \\

\midrule

\textbf{All} & 67.80 & 67.90 & 65.40 & 58.20 & 67.90 & 68.00 & 65.50 & 53.90 \\
\midrule  
\textbf{Punnet} & 98.10 & 100.0 & 99.50 & 98.50 & 98.10 & 100.0 & 99.50 & 78.70 \\
\textbf{OK (Grade 1)} & 66.90 & 71.20 & 73.80 & 63.10 & 66.90 & 71.20 & 73.80 & 61.30 \\
\textbf{Dark (Grade 2)} & 79.30 & 93.50 & 91.90 & 74.50 & 79.70 & 94.10 & 91.00 & 74.80 \\
\textbf{Light (Grade 3)} & 41.60 & 84.10 & 63.90 & 57.10 & 41.60 & 84.10 & 63.90 & 56.30 \\
\textbf{Second (Grade 4)} & 100.0 & 20.90 & 40.90 & 36.10 & 100.0 & 20.90 & 40.90 & 33.10 \\
\textbf{Waste (Grade 5)} & 21.00 & 37.50 & 23.70 & 19.60 & 21.00 & 37.50 & 23.70 & 18.80 \\

\bottomrule
\end{tabular}
\label{tab:mvworks}
\label{table3}
\end{table*}
\normalsize

Although the obtained scores are plausible, given the low number of samples of some classes, we made another experiment by setting the weights of both Grade 4 and Grade 5 to 10, and we set the weight of the classification component in the YOLOv8 loss to 7 as in the previous experiment, the results are reported in Table.~\ref{table4}. One can observe that the overall mAP remains roughly unchanged. While Grade 3 and Grade 5 report some decline int he scores, Grade 1 and Grade 2 show slight improvements, and  Grade 4 shows an interesting improvement. 

\begin{table*}
\captionsetup{size=footnotesize}
\caption{Raspberry instance segmentation and detection scores (\%) obtained by setting the Second (Grade 4) and the Waste (Grade 5) loss weights to 10 and the YOLOv8 classification loss weight to 7. Box metrics stand for the detection task, and Mask metrics refer to the instance segmentation task.}
\setlength\tabcolsep{3pt} 
\centering
\smallskip 
\begin{tabular}{lcccccccc}
\toprule

\textbf{Class} & \textbf{Pre(Box)} & \textbf{Rec(Box)} & \textbf{mAP50(Box)} & \textbf{mAP50-95(Box)} & \textbf{Pre(Mask)} & \textbf{Rec(Mask)} & \textbf{mAP50(Mask)} & \textbf{mAP50-95(Mask)} \\

\midrule

\textbf{All} & 58.80 & 67.20 & 65.00 & 58.10 & 58.80 & 67.30 & 65.10 & 54.20 \\
\midrule  
\textbf{Punnet} & 98.40 & 100.0 & 99.50 & 98.80 & 98.40 & 100.0 & 99.50 & 81.40 \\
\textbf{OK (Grade 1)} & 65.30 & 72.70 & 75.00 & 64.40 & 65.30 & 72.60 & 75.00 & 61.70 \\
\textbf{Dark (Grade 2)} & 83.70 & 92.80 & 92.00 & 77.30 & 83.90 & 93.00 & 92.20 & 76.30 \\
\textbf{Light (Grade 3)} & 54.90 & 69.20 & 59.40 & 53.00 & 54.90 & 69.10 & 59.40 & 54.00 \\
\textbf{Second (Grade 4)} & 33.70 & 50.00 & 45.80 & 38.60 & 33.80 & 50.00 & 45.80 & 36.40 \\
\textbf{Waste (Grade 5)} & 16.80 & 18.80 & 18.50 & 16.30 & 16.80 & 18.80 & 18.50 & 15.20 \\

\bottomrule
\end{tabular}
\label{table4}
\end{table*}
\normalsize

Overall, in terms of balance among the raspberry classes, one may come to the conclusion that the scores reported in Table.~\ref{table3} seem to be the most prominent. In this respect, although the overall mAP remains somewhat limited (65.50\%) due to the challenging nature of the classification task in view of the similarities in color between some classes, we believe that the results can be improvement tangibly by incorporating more training data, especially regarding Grade 3, Grade 4 and Grade 5.

\begin{figure*}
    \centering
    \begin{tabular}{@{}cc@{}} 

        \includegraphics[width=1\textwidth]{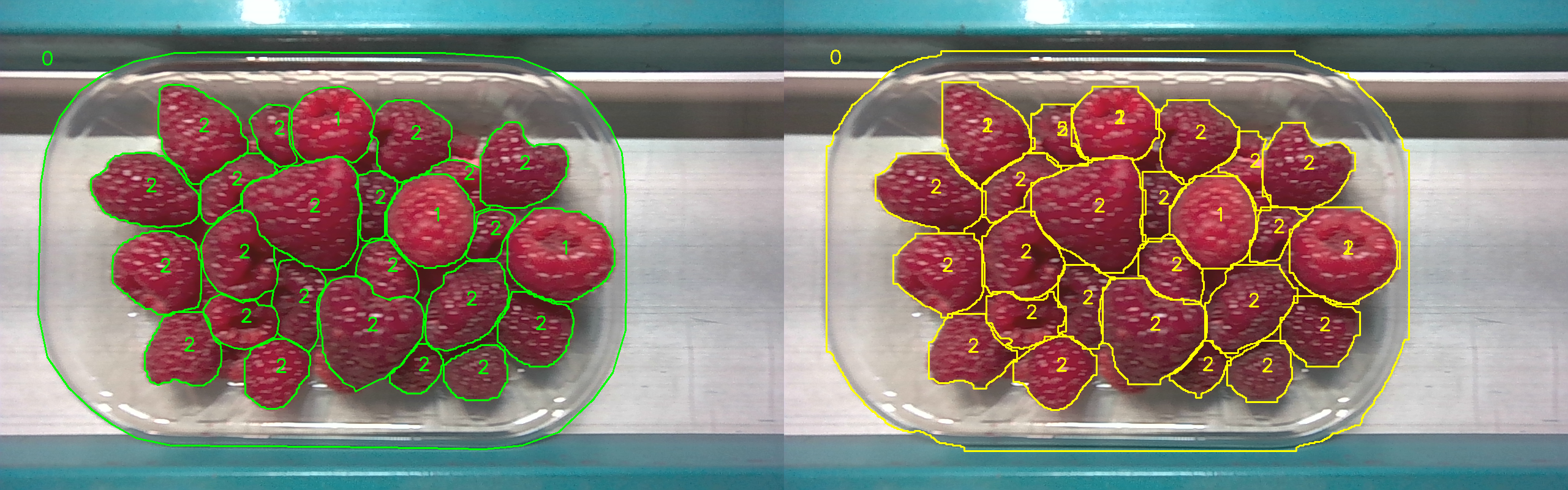} \\

        \includegraphics[width=1\textwidth]{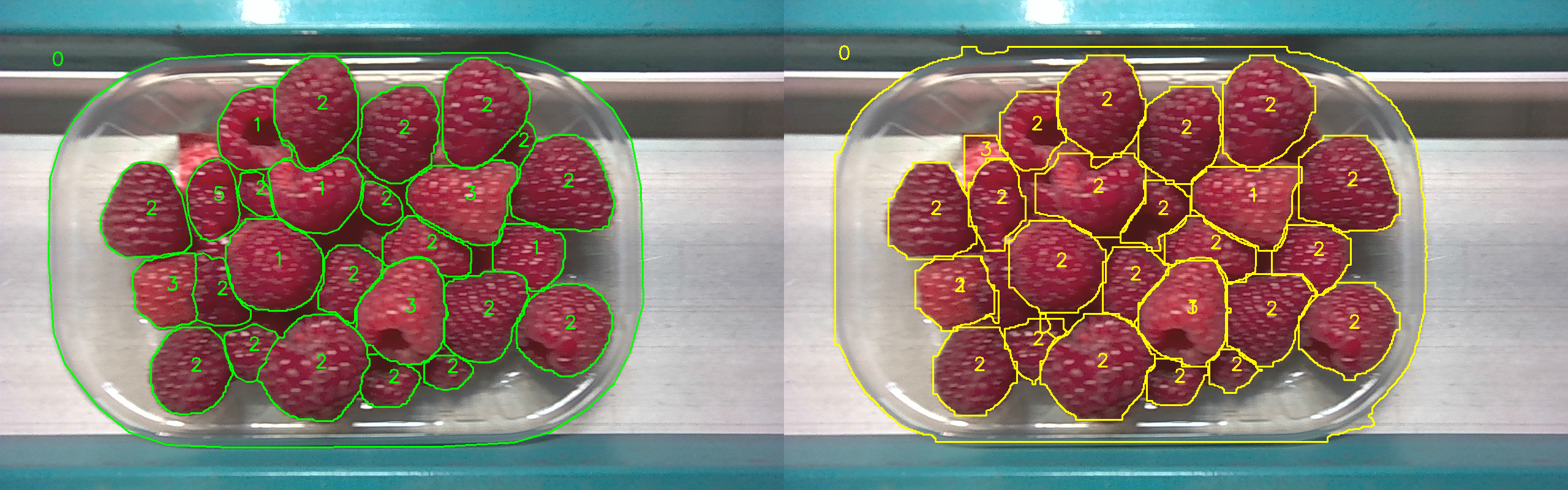} \\
        
        \includegraphics[width=1\textwidth]{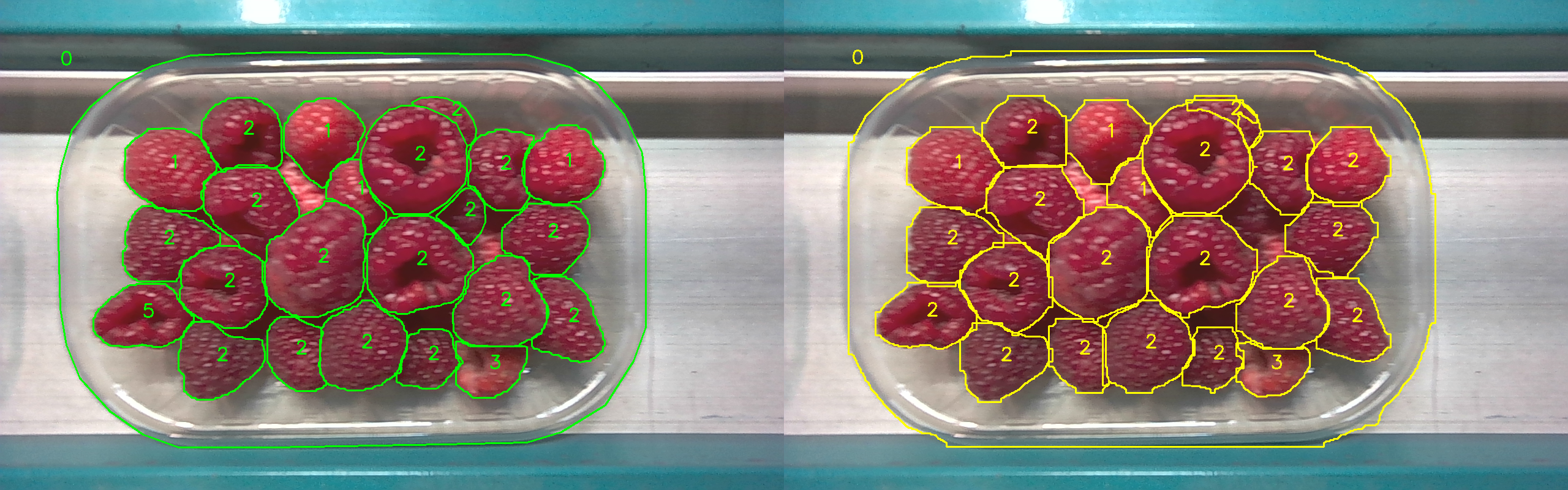} \\
        
    \end{tabular}
    
\caption{Qualitative raspberry segmentation examples. On the left column, the (green) groundtruth annotations of the segments of each raspberry and punnet in the images and their class labels (at the center of the segment for each fruit, and at the top-left corner for the punnet. 0 label refers to the punnet, whilst labels from 1 to 5 indicate gradually the raspberry grades from 1 to 5, respectively). On the right column, the same images with the (yellow) predicted raspberry segments and their class labels.}
\label{qual1}
    
\end{figure*}

We provide several qualitative examples in Fig.~\ref{qual1}. On the left column, the groundtruth annotations of the segments of each raspberry and punnet in the images and their class labels (at the center of the segment for each fruit, and at the top-left corner for the punnet. 0 label refers to the punnet, whilst labels from 1 to 5 indicate gradually the raspberry grades from 1 to 5, respectively). On the right column, the same images with the predicted raspberry segments and their class labels.

\section{Conclusion} \label{conclusions} This paper conducted an early-stage investigation into real-time five-grade raspberry grading using RGB imaging and deep learning-based instance segmentation, a novel problem in automated fruit grading research given the challenges of small, occluded fruits in industrial settings. Our initial results highlighted the impact of training data distribution on grading scores. Our immediate next step is to expand the dataset, particularly for under-represented classes. Future work will explore advanced instance segmentation architectures, enhance robustness to industrial environment variations, and consider multi-modal sensing to improve grading accuracy, supported by the public release of our annotated dataset.
\section{MATCH \& CONTRIBUTION}
This contribution aligns well with the theme of the "ICE IEEE/ITMC 2025 International Conference on Engineering, Technology, and Innovation", part of the IEEE Technology and Engineering Management Society (TEMS). 
The paper explores the application of an AI-driven technology, deep learning-based computer vision, to transform a traditional industrial process – fruit grading – by introducing automation and real-time capabilities. By developing a system for automated raspberry grading, this research emphasizes the importance of leveraging advanced technologies to drive efficiency and innovation in the agricultural sector. The development of a data-driven engineering approach, utilizing a novel dataset and a state-of-the-art deep learning model (YOLOv8), showcases how intelligent systems can respond to the demands for improved quality and throughput in food processing. This contribution addresses the conference's focus on the design of products and services, and adoptions of digital technologies and methods supporting the process of organizational digital transformation in society and the economy.

\section{ACKNOWLEDGEMENTS}

This paper is supported by European Union’s Horizon Europe research and innovation programme under grant agreement No 101092043,
project AGILEHAND (Smart Grading, Handling and Packaging Solutions for Soft and Deformable Products in Agile and Reconfigurable Lines.

\FloatBarrier
\bibliographystyle{IEEEtran}
\bibliography{references.bib}

\end{document}